\documentclass[letterpaper, 10 pt, conference]{ieeeconf}  

\IEEEoverridecommandlockouts                              

\usepackage{graphics} 
\usepackage{epsfig} 
\usepackage{mathptmx} 
\usepackage{mathrsfs}
\usepackage{times} 
\usepackage{amsmath} 

\usepackage{amsthm}
\usepackage{amssymb}  
\usepackage{wasysym}
\usepackage{txfonts}
\usepackage{bbm}
\usepackage{color}  
\usepackage[dvipsnames]{xcolor}
\usepackage{indentfirst}
\usepackage{multirow}

\usepackage{perl_acronyms}
\usepackage{perl_misc}

\usepackage{color, colortbl}
\usepackage{algorithm}
\usepackage{algpseudocode}

\usepackage{tikz}
\usetikzlibrary{arrows,shapes,trees,calc}
\usetikzlibrary{backgrounds}
\usepackage{pgfplots}
\pgfplotsset{compat=1.14}
\usepgfplotslibrary{polar}
\usepgfplotslibrary{patchplots}

\newcolumntype{P}[1]{>{\centering\arraybackslash}p{#1}}
\usepackage{boldline}
 \usepackage{textcomp}
\newcounter{RamCount}
\newcounter{TaliaCount}
\newcounter{DanCount}
\newcounter{ShannonCount}
\newcounter{MaggieCount}

\usepackage[
    style=ieee,
    doi=false,
    isbn=false,
    url=false,
    eprint=false,
    backend=bibtex,
    natbib=true
    ]{biblatex}

\bibliography{references}

\definecolor{Gray}{gray}{0.9}
\newcolumntype{g}{>{\columncolor{Gray}}c}

\usepackage{hyperref}

\title{\LARGE \bf Emulating duration and curvature of coral snake anti-predator thrashing behaviors using a soft-robotic platform}
\author{Shannon M. Danforth$^{1*}$, Margaret Kohler$^{2*}$,  Daniel Bruder$^{3*}$, \\
Alison R. Davis Rabosky$^{4}$, Sridhar Kota$^{5}$, Ram Vasudevan$^{6}$, Talia Y. Moore$^{7}$ 
\thanks{$^{*}$ indicates equal contribution to this project.}
\thanks{This material is based upon work supported by the National Science Foundation Graduate Research Fellowships awarded to SMD and DB under Grant No. 1256260 DGE. 
Additional support for this material was provided by University of Michigan MCubed Award 1499 to RV, ARDR, DB, and TYM.}%
\thanks{$^{1,2,3,5,6}$ Mechanical Engineering, University of Michigan, Ann Arbor, MI.
        {\tt\small \{sdanfort, mjkohler, bruderd, kota, ramv\}@umich.edu}}
\thanks{$^{4}$Ecology and Evolutionary Biology, Museum of Zoology, University of Michigan, Ann Arbor, MI. {\tt\small ardr@umich.edu}}%
\thanks{$^{6,7}$Robotics Institute, University of Michigan, Ann Arbor, MI. {\tt\small taliaym@umich.edu}}%
}

\begin{document}
\setlength{\textfloatsep}{8pt}

\maketitle
\thispagestyle{empty}
\pagestyle{plain}

\begin{abstract}
This paper presents a soft-robotic platform for exploring the ecological relevance of non-locomotory movements via animal-robot interactions.
Coral snakes (genus \emph{Micrurus}) and their mimics use vigorous, non-locomotory, and arrhythmic thrashing to deter predation.
There is variation across snake species in the duration and curvature of anti-predator thrashes, and it is unclear how these aspects of motion interact to contribute to snake survival.
In this work, soft robots composed of fiber-reinforced elastomeric enclosures (FREEs) are developed to emulate the anti-predator behaviors of three genera of snake.
Curvature and duration of motion are estimated for both live snakes and robots, providing a quantitative assessment of the robots' ability to emulate snake poses.
The curvature values of the fabricated soft-robotic head, midsection, and tail segments are found to overlap with those exhibited by live snakes. 
Soft robot motion durations were less than or equal to those of snakes for all three genera.
Additionally, combinations of segments were selected to emulate three specific snake genera with distinct anti-predatory behavior, producing curvature values that aligned well with live snake observations.
\end{abstract}

\begin{figure}
    \centering
    \includegraphics[width=\columnwidth]{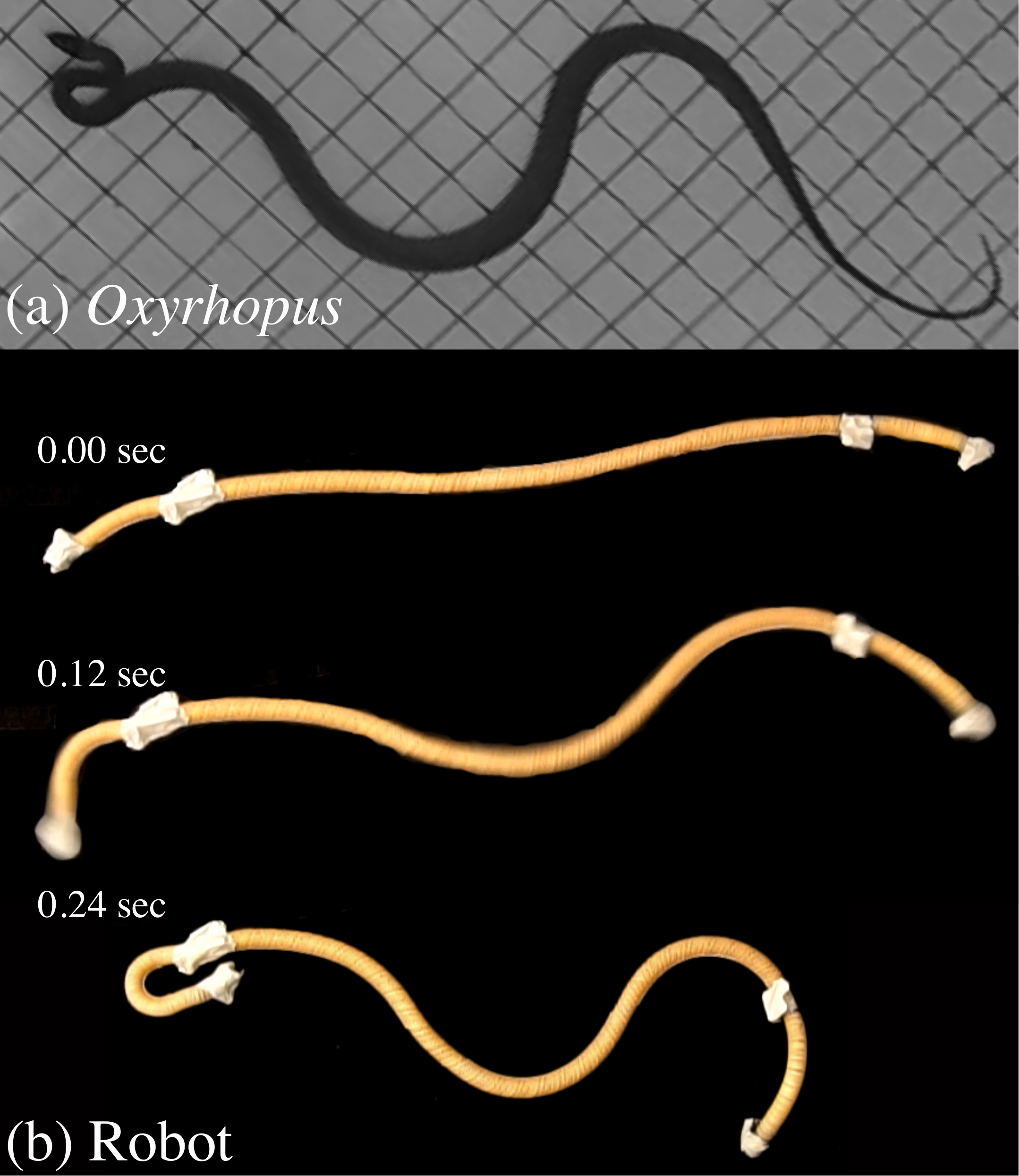}
    \caption{This is an overview figure that shows a typical a) \emph{Oxyrhopus} snake anti-predator posture and b) the soft robot designed to emulate this genus of snake.
    The robot video was captured at 60 fps and the blur at 0.12 sec is due to rapid motion.}
    \label{fig:snakevrobot}
\end{figure}

\section{Introduction}\label{sec:introduction}
Soft robots have the potential to revolutionize human-robot interaction due to their biocompatible materials, variable stiffness, and independence from electronic actuation \cite{rus2015design, laschi2016}.
Unfortunately, current soft robots fall short of these lofty goals, which still require significant advances in control, actuation, and manufacturing \cite{kim2013soft, majidi2014soft}.
With present capabilities, however, soft robot technology can already be used for less-demanding applications.
In this work, we demonstrate its use for enhancing biological inquiry into non-locomotory movement by enabling more realistic robot-animal interaction experiments than are possible with rigid biomimetic robots.

Soft robot design has been greatly inspired by the locomotion of limbless organisms, such as snakes.
For some snakes, non-locomotory motion is equally important for survival \cite{brodie1992}.
Coral snakes (genus \emph{Micrurus}) are venomous snakes that signal their lethal toxicity to predators with colorful patterns and vigorous thrashing displays \cite{Moore2019,RodriguesFranca2016,DuVal2006}.
These non-locomotory anti-predator behaviors are so effective that several groups of non-venomous snakes benefit from partially mimicking these color patterns and behaviors \cite{greene1979}.
While the effect of color pattern has been tested using painted stationary models, precise control of movement is necessary to test for an effect on deterring predation \cite{Brodie1993}.
Robots developed to precisely emulate biological motions have been used for hypothesis testing \cite{libby2012, patricelli2009}, but few can be safely hunted by live predators.
Soft robots have the advantage of being generally safer for interaction with live animals, and some have been designed with motion reminiscent of biological organisms \cite{wehner2016,Suzumori2006,Laschi2012,Ren2019}.
Therefore, soft robots that can be safely destroyed and are designed to precisely emulate biological motion would be the ideal choice for measuring the effectiveness of various  anti-predator behaviors.

Fiber-reinforced elastomeric enclosures (FREEs) are a class of soft, fluid-driven actuators composed of fibers wound around elastomer tubes.
The fibers enforce a volumetric constraint, which imposes specific deformations such as extension, torsion, bending, and coiling when the tube is pressurized. 
These robots can act as safe and ethical prey items for live predators because they are pneumatically actuated and made from biocompatible soft materials.

In this work, we develop soft robots composed of FREEs that can precisely mimic the thrashing vigor and curvature exhibited by multiple species of snake while safely interacting with live animal predators (Fig. \ref{fig:snakevrobot}).
Our contributions are as follows.
First, a framework for the design and fabrication of such robots.
Second, the first quantitative comparison of the anti-predatory thrashing behavior between \emph{Micrurus} coral snakes and two of its mimics.
Third, an approach for comparing non-locomotory behavior between bio-inspired robots and their biological templates.

The three snake genera used for comparison in this study are described in Section \ref{sec:sayhellotosnakes}.
The theoretical justification for our method of inducing curvature in a FREE as well as a description of the fabrication method is provided in Section~\ref{sec:theory}.
The study methods, including the live snake observations, robot experiment, and curvature estimation, are discussed in Section \ref{sec:methods}.
Comparisons of motion and curvature between live snake and robot are presented and discussed in Section \ref{sec:results}.
Concluding remarks are provided in Section~\ref{sec:conclusion}.

\section{Biological Template}\label{sec:sayhellotosnakes}
\begin{table}
\centering
\caption{A limited number of pieces can be assembled in multiple ways to emulate three distinct genera of snake.}
\renewcommand\arraystretch{1.5}
\begin{tabular}{|c|P{1.5cm}|P{1.5cm}|P{1.5cm}|}
     \hline
     \rowcolor{gray!25} Genus & Head & Midsection & Tail \\
     \hline
     \multirow{2}{*}{\emph{Atractus}} & \multirow{2}{*}{Straight} & U-Curve S-Curve & \multirow{2}{*}{Coil} \\
     \hline
     \emph{Micrurus} & Kink & S-Curve & Coil \\
     \hline
     \emph{Oxyrhopus} & Kink & S-Curve & Straight \\
     \hline
\end{tabular}
\label{table:shape2species}
\end{table}
This section describes the behaviors of the three snake genera that serve as the template for soft-robotic design.
We captured snakes in funnel traps and opportunistically during transects in the Amazonian rainforests of Peru in November and December 2018 (see \cite{Moore2019} for details). 
Researchers were equipped with knee-high rubber boots, snake hooks, and venom defender gloves when handling potentially venomous snakes.
Within 24 hours of capture, snakes were released, one at a time, in a pop-up corrugated plastic behavioral arena marked with gridlines as visual fiduciary markers.
Snakes were video-recorded at 120fps with two Hero 4 Black (GoPro) cameras as they were exposed to stimuli to simulate predators.
One one-minute long trial of anti-predator behavior was selected for each snake genus.

All snakes displayed a vigorous, non-locomotory `thrashing' behavior, consisting of rapidly straightening and curling the body interspersed by periods of immobility.
\emph{Micrurus lemniscatus} was the most common species of coral snake in our field-collected dataset.
These snakes thrash with distinctive lateral kinks in the anterior portion of the head, lateral curves of the body, and an elevated, coiled tail.
\emph{Atractus elaps} is distantly related to the coral snake \cite{davisrabosky2016} and thrashes with few lateral curves of the body and an elevated, coiled tail when threatened.
The anterior portion of the snake remains unkinked.
\emph{Oxyrhopus melanogenys} was the most common mimic species in our dataset.
These snakes thrash with distinctive lateral kinks in the anterior portion of the head and lateral curves of the body.
The posterior portion of the snake remains uncoiled.
In the observations used for this work, 19 (\emph{Oxyrhopus}) to 41 (\emph{Micrurus}) thrashes were exhibited in a given trial.

The key properties of the snake behaviors we seek to emulate are the rapid thrash durations and the distinct regions of curvature along the length of the body during the periods of rest immediately post-thrash.
The duration of each thrash was recorded by counting the number of frames in which motion-blur was observed.

Because the anti-predator behavior of mimic snakes includes only a subset of the behaviors demonstrated by the \emph{Micrurus} coral snake \cite{greene1979}, a limited range of robotic motion-primitives can be mixed and matched to mimic the behaviors of several snake species (Table \ref{table:shape2species}).

\section{Design and Fabrication of Snake Robots}\label{sec:theory}
\begin{figure}
    \centering
    \begin{tikzpicture}
        \node[style={anchor=center}] (image) {\includegraphics[width=0.8\linewidth]{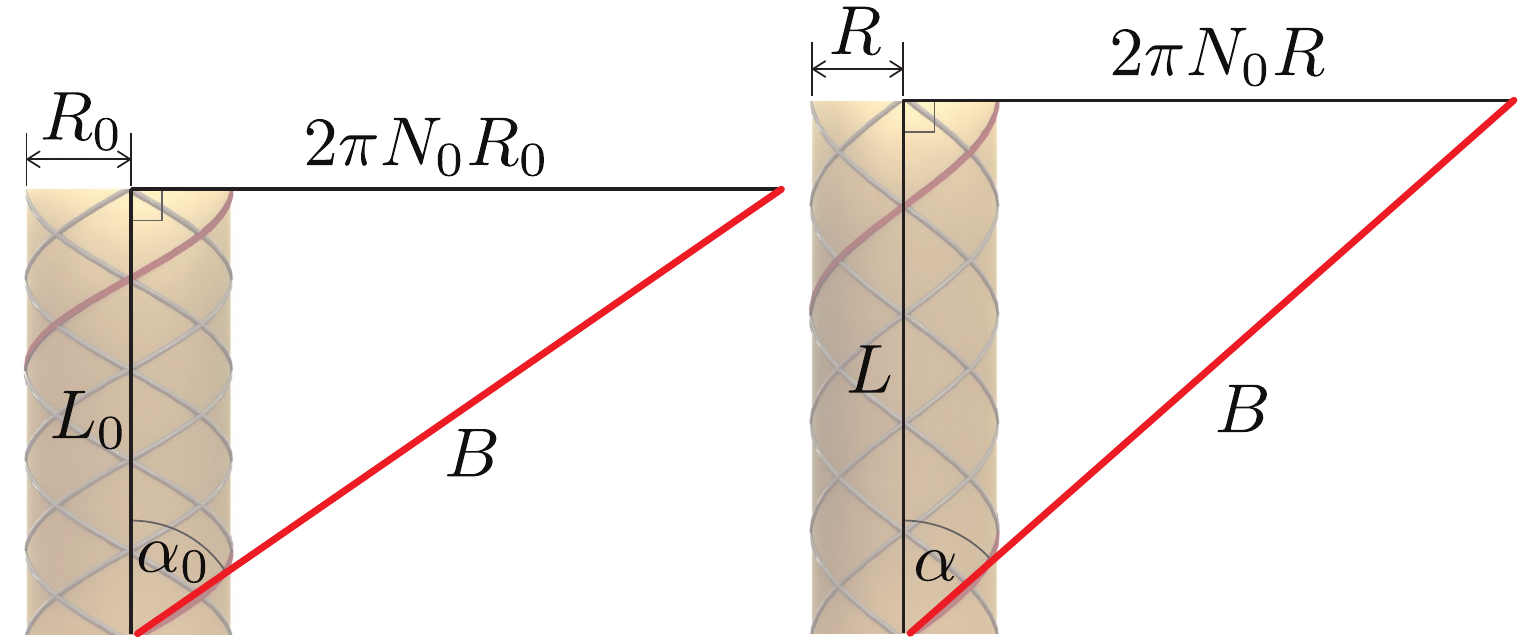}};
    \end{tikzpicture}
    \caption{The shape of a FREE is parametrized by its length $L$, radius $R$, and fiber angle $\alpha$. 
    These parameters are geometrically related via the right triangle formed by unwinding a fiber and laying it flat.
    The parameters for a FREE in its relaxed configuration are denoted $L_0$,  $R_0$, $\alpha_0$, and the fiber length is denoted $B$ (left).
    When pressurized, a FREE increases in volume but the length of its fibers remains constant (right). For FREEs with $\alpha_0 > 54.74^\circ$, this change in volume causes the length to increase, the radius to decrease, and the fiber angle to move closer to $54.74^\circ$.}
    \label{fig:fiber-unrolled}
\end{figure}
FREEs have been shown to be capable of achieving many types of deformation, including contraction/extension \cite{tondu2012modelling}, twist \cite{bruder2017model}, screw \cite{krishnan2015kinematics}, bending \cite{bishop2015design, polygerinos2015modeling}, and coiling \cite{bishop2013towards}.
These deformations result from the geometric constraints imposed by the fiber reinforcements under volumetric expansion.
In this work, customizable bending segments are created by adhering an additional strain-limiting fiber along one side of an extending-type FREE.
This section provides the theoretical justification for this method of inducing curvature, describes how curvature is influenced by the angles at which fiber reinforcements are wrapped, and describes our fabrication methods.

\subsection{Extending-type FREEs}
FREEs that have two families of fibers wrapped at equal and opposite winding angles are known colloquially as McKibben artificial muscles or pneumatic artificial muscles (PAMs).
Using the fiber angle labeling convention shown in Fig. \ref{fig:fiber-unrolled}, FREEs with fiber angle of magnitude less than $\arctan{ \sqrt{2} } \approx 54.74^\circ$ will contract under an increase in volume, while those with fiber angle greater than $54.74^\circ$ will extend \cite{tondu2012modelling}.
Therefore, the most extreme deformation of a FREE occurs at the configuration in which the fibers achieve this so-called ``neutral angle'', and no further contraction or extension is achievable beyond this point.

An ideal cylindrical FREE in its relaxed configuration (i.e. when internal gauge pressure is zero and no external loads are applied) can be described by a set of three parameters, $L_0$, $R_0$, and $\alpha_0$, where $L_0$ represents the relaxed length of the FREE, $R_0$ represents the relaxed radius, and $\alpha_0$ the fiber angle (see Fig.~\ref{fig:fiber-unrolled}). 
For notational convenience, we define two other quantities from these parameters,
\begin{align}
	B &= \frac{L_0}{\cos{\alpha_0}} \\
	N &= \frac{L_0}{2 \pi R_0} \tan{\alpha_0}
\end{align}
where $B$ is the length of one of the FREE fibers and $N$ is the total number of revolutions the fiber makes around the FREE in the relaxed configuration \cite{bruder2018force}.

When pressurized, a FREE increases in volume, which results in changes to its length and radius.
Under the common assumption that the reinforcement fibers are inextensible, the FREE's length and radius are coupled.
To illustrate why, consider a single fiber that is unrolled so that it lays flat and forms a right triangle with height $L$ and hypotenuse $B$, as shown in Fig. \ref{fig:fiber-unrolled}.
As the volume of the cylinder increases, its length and radius change, but the length of the fiber does not.
As a consequence, an expression for the radius $R$ in terms of the length $L$ can be derived by applying the Pythagorean Theorem to the right triangle shown in Fig.~\ref{fig:fiber-unrolled}.
\begin{align}
	R &= \frac{ \sqrt{ B^2 - L^2 } }{ 2 \pi N} . \label{eq:r}
\end{align}

In a similar fashion, the theoretical maximum length for a cylindrical FREE with initial fiber angle $\alpha > 54.74^\circ$ can be calculated using trigonometry.
As previously stated, the maximum length $L_\text{max}$ occurs when a FREE has been deformed to the point where its fiber angle equals the ``neutral angle'' of $\approx 54.74^\circ$.
As pictured in Fig.~\ref{fig:fiber-unrolled}, this length corresponds to the 
height of a triangle with hypotenuse $B$ and angle $54.74^\circ$, which is given by the following expression,
\begin{align}
    L_\text{max} &= B \, \cos{54.74^\circ} = L_0 \, \frac{\cos{54.74^\circ}}{\cos{\alpha_0}}.
\end{align}

\subsection{Converting Extension to Bending}
\begin{figure}
    \centering
    \begin{tikzpicture}
        \node[style={anchor=center}] (image) {\includegraphics[width=0.7\linewidth]{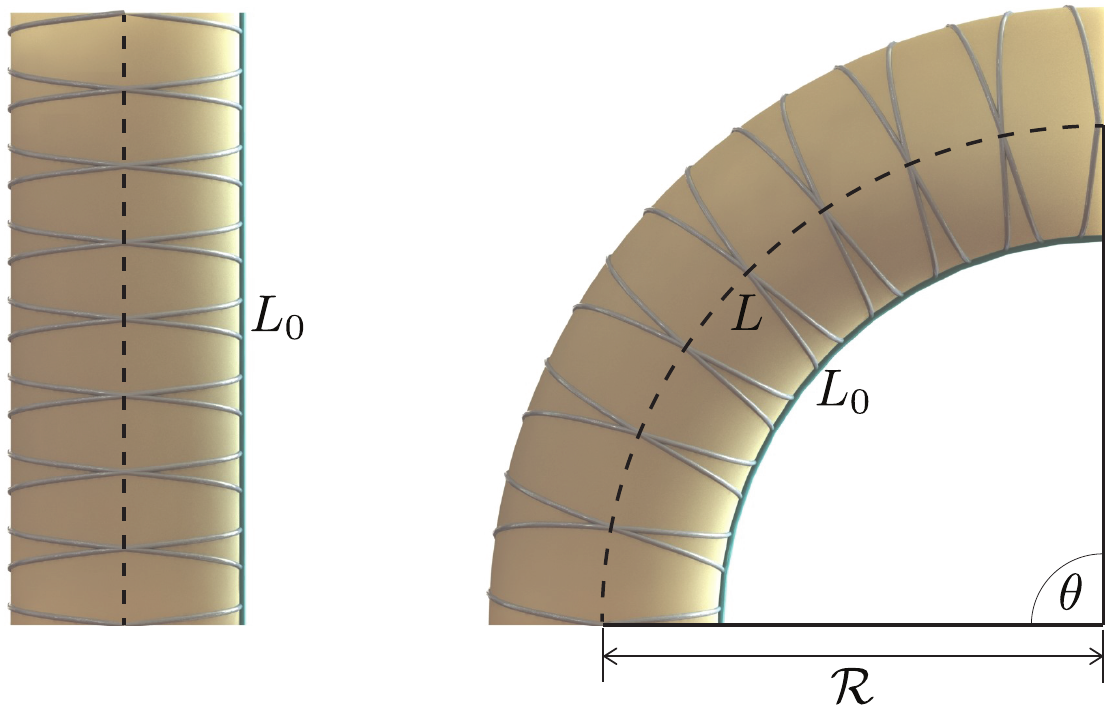}};
    \end{tikzpicture}
    \caption{The additional fiber of length $L_0$ adhered to an extending-type FREE permits no extension along its right side.
    As the volume of the FREE increases from its relaxed volume (left) to a larger volume (right), so does the arc length along its central axis $L$.
    However, the arc length along its right side remains constant, which induces a curvature of radius $\mathcal{R}$ and angle $\theta$.}
    \label{fig:straight2curve}
\end{figure}

Extension can be converted into bending by adding a strain-limiting fiber along the body of a FREE, as shown in Fig. \ref{fig:straight2curve}.
As the volume of the FREE increases, the fiber permits no extension of one of its sides, which results in a bent geometry.
This bending can be quantified in terms of a radius of curvature $\mathcal{R}$ and angle of curvature $\theta$, as shown in Fig.~\ref{fig:straight2curve}.
Assuming that the arc length of the central axis of a bent FREE is equal to the extended length of an identical FREE without the strain-limiting fiber and that the inner arc length is equal to $L_0$, the following expressions hold true:
\begin{align}
    L &= \mathcal{R} \theta,    \label{eq:LRtheta} \\
    L_0 &= ( \mathcal{R} - R ) \theta.  \label{eq:L0Rrtheta}
\end{align}
Combining Equations \eqref{eq:LRtheta} and \eqref{eq:L0Rrtheta}, solving for $\mathcal{R}$, and taking its reciprocal yields an expression for the curvature of the FREE, denoted $\mathcal{K}$,
\begin{align}
    \mathcal{K} &= \frac{1}{\mathcal{R}} = 2 \pi N \left( 1 - \frac{L_0}{L} \right) \left( B^2 - L^2 \right)^{-\frac{1}{2}}.
    \label{eq:curvature}
\end{align}
The maximum curvature $\mathcal{K}_\text{max}$ is given by substituting $L_\text{max}$ into Equation \ref{eq:curvature}, with the following result,
\begin{align}
    \mathcal{K}_\text{max} &= \frac{1}{R_0} \left( \frac{\sin{\alpha_0}}{\sin{54.74^\circ}} \right) \left( 1 - \frac{\cos{\alpha_0}}{\cos{54.74^\circ}} \right) .
    \label{eq:Kmax}
\end{align}
The accuracy of this curvature prediction is limited by the model's simplifying assumptions and unmodeled effects such as elasticity and friction.
However, this model does capture a qualitative dependence between fiber angle and curvature that is useful for informing design.
One can verify that the expression for $\mathcal{K}_\text{max}$ is monotonically increasing with respect to $\alpha_0 \in [ 54.74^\circ , 90^\circ )$.
This relationship implies that for two bending FREEs with the same relaxed radius $R_0$, if $\alpha_{0_1} < \alpha_{0_2}$ then $\mathcal{K}_{\text{max}_1} < \mathcal{K}_{\text{max}_2}$.
In other words, the FREE with the larger relaxed fiber angle has a larger maximum curvature.

\subsection{Fabrication of Snake Robots}

We create soft robots that imitate the curvature and thrashing motions of snakes upon pressurization by connecting several bending FREEs in series.
To achieve desired shapes, we tune the curvature of each bending segment by modifying the relaxed fiber angle according to the relationship captured by \eqref{eq:Kmax}.
For sections of the snake body that require the largest curvature, we utilize FREEs with large fiber angles (e.g. $\alpha_0 \in [80^\circ , 90^\circ) $) and for sections that require less curvature we use smaller fiber angles (e.g. $ \alpha \in [60^\circ , 70^\circ] $).

Bending segments are fabricated by winding cotton fibers around $0.95$ cm ($3/8 "$) inner diameter latex tubing then dipping them in Monster liquid latex (Kangaroo Manufacturing, Tempe, AZ) to adhere the fibers to the tube.
Fiber winding is automated through the use of a modified X-Winder machine (X-Winder, Cincinnati, OH).
After the liquid latex has fully cured, bending segments are trimmed to desired lengths and connected in series using two-way barbed tube connectors and hose clamps.
A single pressure line connected to an air compressor is used to simultaneously pressurize all bending segments of the fully assembled snake robot.

For a straight head, circumferential fibers are wound at 67.5\textdegree{}.
To emulate the kinked necks, circumferential fibers are wound at 89\textdegree{} and a strain-limiting fiber is placed along the side of the tube to create the concave portion of each kink when inflated.
Lateral curves along the body are achieved by winding circumferential fibers at 67.5\textdegree{} and adding long strain-limiting fibers along opposite sides of the tube where the concave portion of each curve will be when inflated.
A straight tail is emulated by wrapping circumferential fibers around a short piece of latex tubing at 89\textdegree{} with no strain-limiting fiber. A curved tail is emulated by attaching one strain-limiting fiber along the length of a short piece of latex tubing wrapped circumferentially at 89\textdegree{}.

\section{Methods}\label{sec:methods}

\subsection{Experiment}\label{sec:experiment}
The assembled FREE tubes were pressurized to 310 kPa (45 psi) by rapidly opening a valve connecting them to a compressed air source.
Each snake-mimicking combination of head, body, and tail was recorded 10 times with a minimum of 5 seconds of rest between actuations.
The unmodified coiling tail segment was tested individually.
Trials were recorded at 60fps using a Sony Alpha 7 III camera, and the duration of actuation was also measured by counting the number of frames in which motion was observed.


\subsection{Curvature Estimation and Analysis}\label{subsec:curvature}
\begin{figure}
    \centering
    \includegraphics[width=0.9\columnwidth]{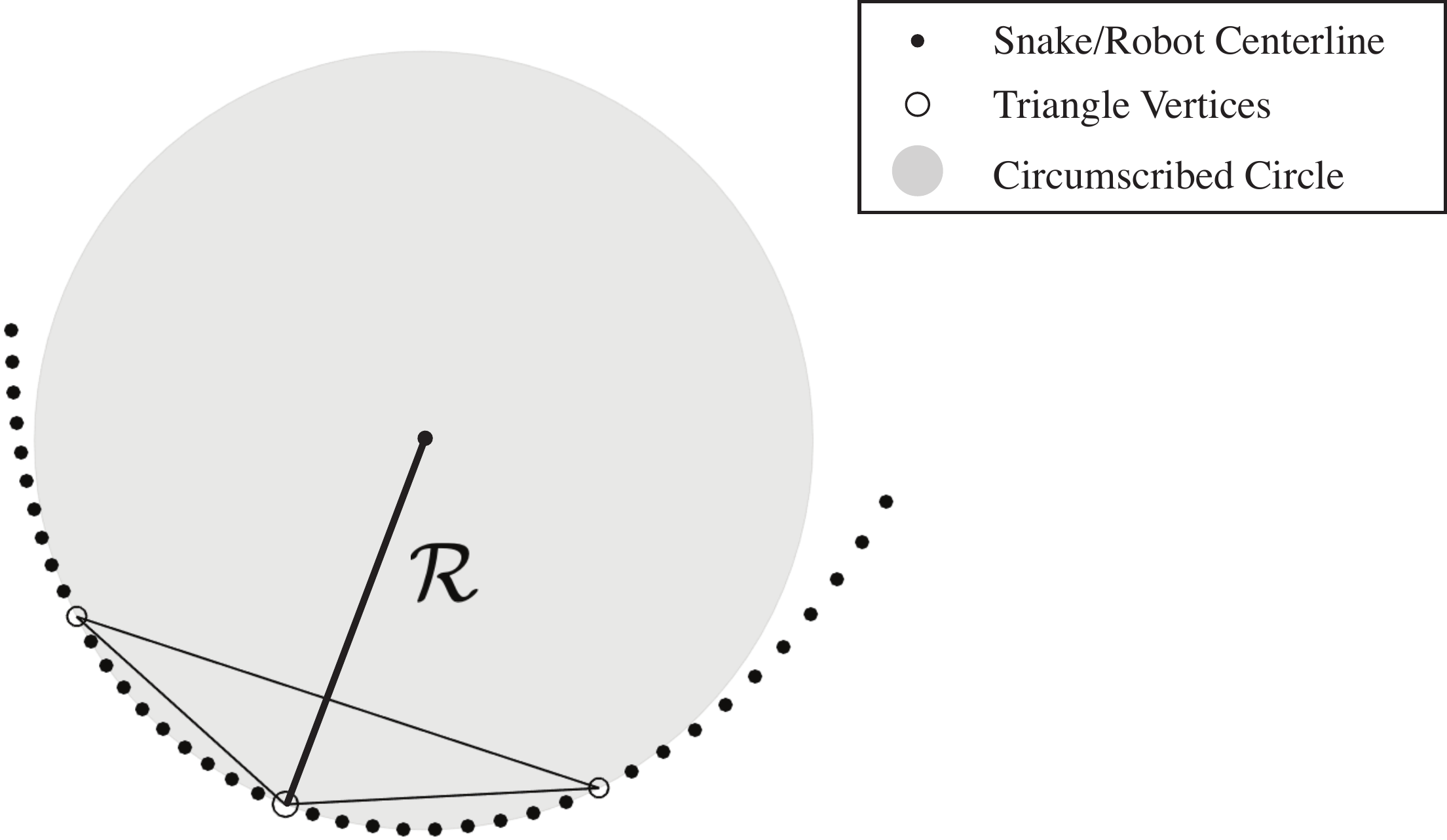}
    \caption{Method for estimating the radius of curvature along each point of the snake or robot centerline (filled black circles). A triangle (thin black line) is formed using the point of interest (larger hollow circle) and the tenth points to the left and right (smaller hollow circles). These polygon vertices form a circumscribed circle (gray), which is used to estimate the radius of curvature $\mathcal{R}$ at the point of interest. }
    \label{fig:curvature_calc}
\end{figure}

For both snakes and robots, we used Quicktime7 to view the videos frame-by-frame.
The frame representing the final snake or robot position after each thrash was selected for computing curvature in a similar method used by \cite{Moore2019}.
We first performed a projective transformation on each post-thrash frame in MATLAB to produce a top view of the snake or robot.

Due to large amounts of self-occlusion in the live snake observations, we used ImageJ to manually trace the centerline of the snake body from the rectified video frames.
Automation of centerline-finding for the robot observations was possible because we designed the robots to prevent self-occlusion while simultaneously achieving areas of high curvature.
We used native MATLAB functions to perform morphological operations on binarized versions of each post-thrash image, reducing pixels to a single centerline along the snake robot.
These points were then automatically ordered from head to tail using knowledge of the experimental setup.
For both snake and robot, we resampled the centerline to produce 500 points, then corrected for noise in centerline tracing by smoothing the centerline using a moving average method with a span of 30 points.

To estimate the radius of curvature $\mathcal{R}$ at each point along the body, we first created a triangle using the point of interest and the tenth points to its left and right.
Next, we constructed a circumscribed circle from the vertices of this triangle \cite{kimberling1998}.
The estimated $\mathcal{R}$ is the distance from the point of interest to the circumcenter of this circle, shown in Fig. \ref{fig:curvature_calc}.
We normalized each $\mathcal{R}$ by the total length of the body centerline.
The reciprocal of these values provides an estimate of the curvature $\mathcal{K}$. 
The radius of curvature was not computed for the first and last ten points along the centerline, as no triangle could be formed with uniform spacing in this case.

\section{Results and Discussion}\label{sec:results}

\begin{figure}[b!]
    \centering
    \includegraphics[width=0.9\columnwidth]{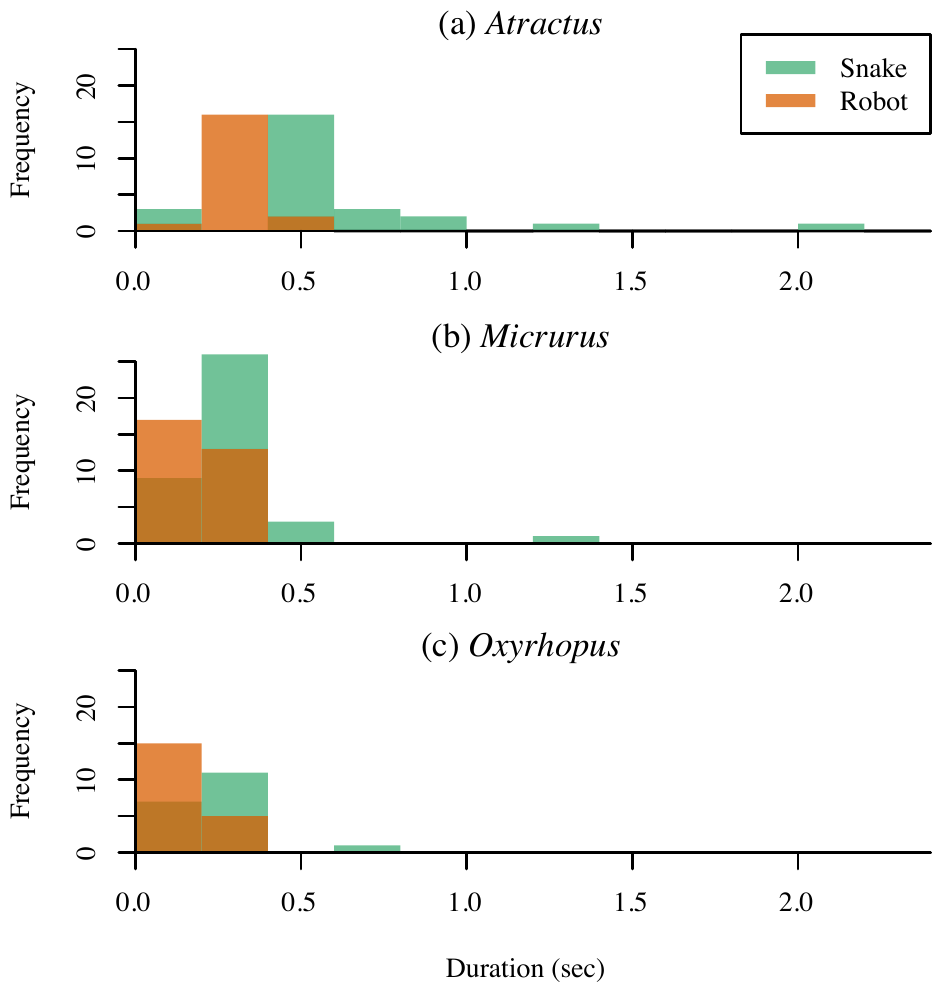}
    \caption{The duration of soft robot thrash is equal to or less than snake thrashing durations.}
    \label{fig:durations}
\end{figure}

\begin{figure*}[t]
    \centering
    \includegraphics[width=0.9\textwidth]{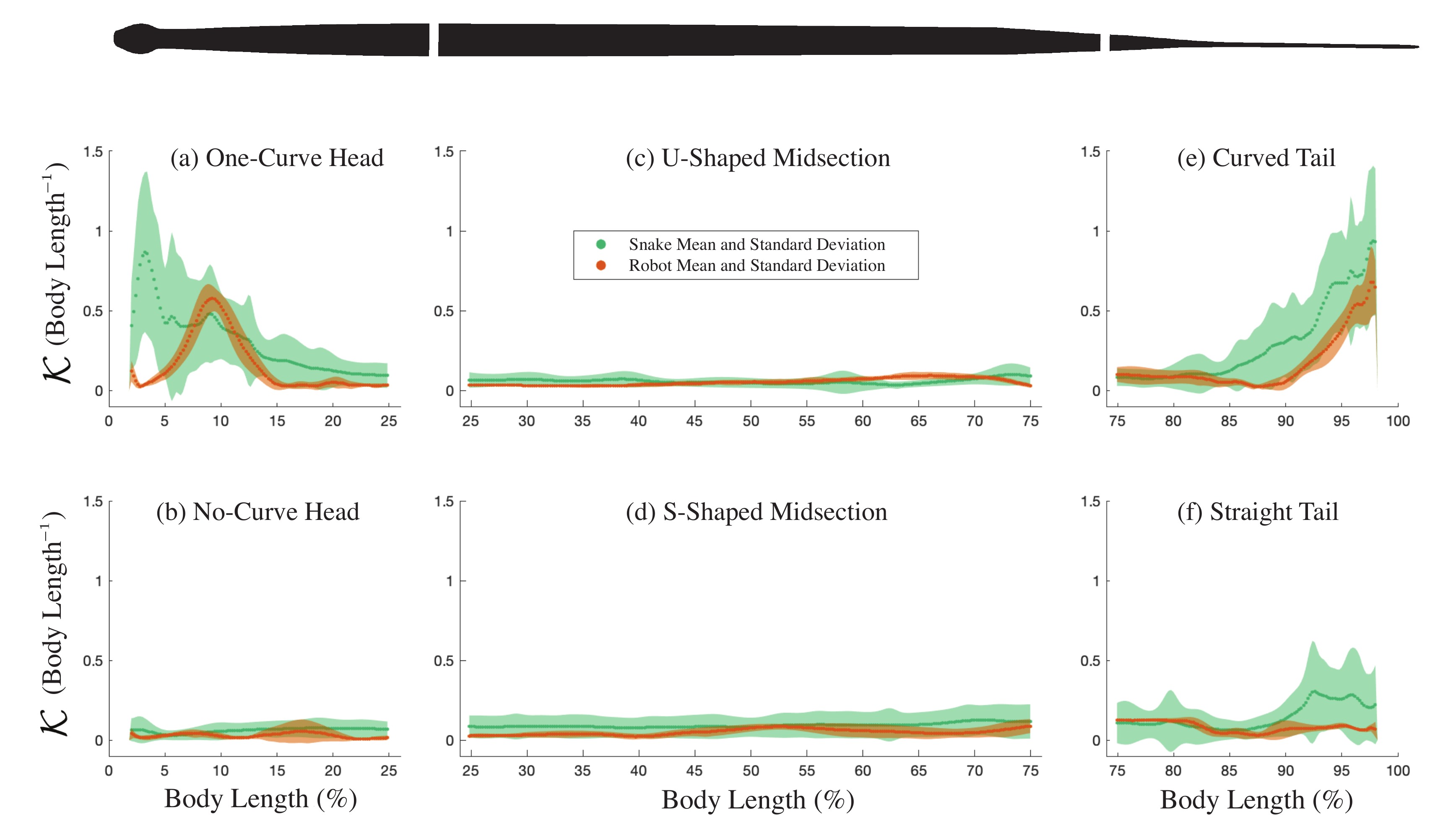}
    \caption{Each modular FREE component has comparable behavior to the snake.
    Snake (green) and robot (orange) heads have (a) lateral curvature or (b) no curvature.
    Snake and robot midsections have (c) U-shaped or (d) S-shaped lateral curvature.
    Snake and robot tails have (e) high curvature and (f) no curvature.}
    \label{fig:head_middle_tail}
\end{figure*}

\subsection{Thrash Durations}\label{sec:durations}
The snake robot motions are similar in duration to the thrashing motions observed in real snakes (Fig.~\ref{fig:durations}), but the robots consistently actuated slightly faster than the real snakes.
In future prototypes, the speed of actuation can be reduced by modifying the tube connecting the compressed air source to the robot.
Decreasing the diameter of this tube introduces more fluid resistance, which increases the time needed for the snake robot to fully inflate.

\subsection{Local Shapes}\label{sec:localshapes}

The curvature mean and standard deviation between robots and live snakes, separated into different shapes for head, midsection, and tail segments, is shown in Fig. \ref{fig:head_middle_tail}.
Robotic head segments were compared to behavior exhibited by the first $25\%$ of snake body length, robotic midsections were compared to the middle $50\%$, and robotic tails were  to the last $25\%$.

Despite the visually close match, the robot mean curvature sometimes dips below the snake standard deviation envelope.
By design, the snake robots exhibited less self-occlusion than the live snakes, and motion was restricted to the 2D plane.
While this design choice made automating the centerline-finding process straightforward, it also prevented the robots from achieving as high of curvatures as the live snakes, which often exhibited sharp head kinks, overlapping midsections, and tail curves in 3D.
The snake robots also exhibited less variation in curvature, which was expected due to a more consistent morphology in robot trials than live snakes.

\subsubsection{Head Segments}\label{sec:headresults}
The robot mean curvature matches closely with that of the second head kink observed in \emph{Micrurus} and \emph{Oxyrhopus}.
The first kink in the live snake, located close to $0\%$ body length, remained constant throughout snake trials.
While a curve of this length and magnitude was not achieved by the robot experiment here, it could emulated by adding a pre-kinked piece to the head of the snake robot.
The straight head segment exhibited by \emph{Atractus} was achieved by the robot, with a larger standard deviation around $15-20\%$ body length resulting from noise in the automated centerline tracing.

\subsubsection{Midsection Segments}\label{sec:bodyresults}
The snakes' S- and U-shaped midsection curves are generally ``softer'' than those in the head and tail segments.
Thus, both shape categories resulted in low curvature magnitudes, which the robot segments were able to emulate.
In U-shaped observations, the robot achieved higher mean curvature than the live snake at $60-70\%$ of the body length, demonstrating that the snake robot midsection is capable of higher deformations.
The S-shaped curvature magnitude is higher in the live snakes than robots, as the snakes occasionally curved their midsection tightly.
These sharper midsection curves often resulted in self-occlusion, which was avoided in the robot observations.

\subsubsection{Tail Segments}\label{sec:tailresults}

The tail segment was divided into straight and curved shapes for analysis. 
Because \emph{Micrurus} exhibited a tightly-coiled tail that remained in 3D for many trials, the coiled portion of the tail was not included in the 2D snake-centerline tracing.
The curved tail data shown in Fig. \ref{fig:head_middle_tail} is from \emph{Atractus} alone.
We are able to achieve similar curvature values for both straight and curved tails with the robots.

\subsection{Species Comparisons}\label{sec:speciesresults}

Intra- and interspecific variation in curvature associated with anti-predator thrashing has only been quantitatively characterized in the \emph{Micrurus} genus \cite{Moore2019}.
This work represents the first quantitative behavioral characterization of harmless snakes that mimic the thrashing behaviors of \emph{Micrurus} snakes.

\begin{figure}
    \centering
    \includegraphics[width=0.9\columnwidth]{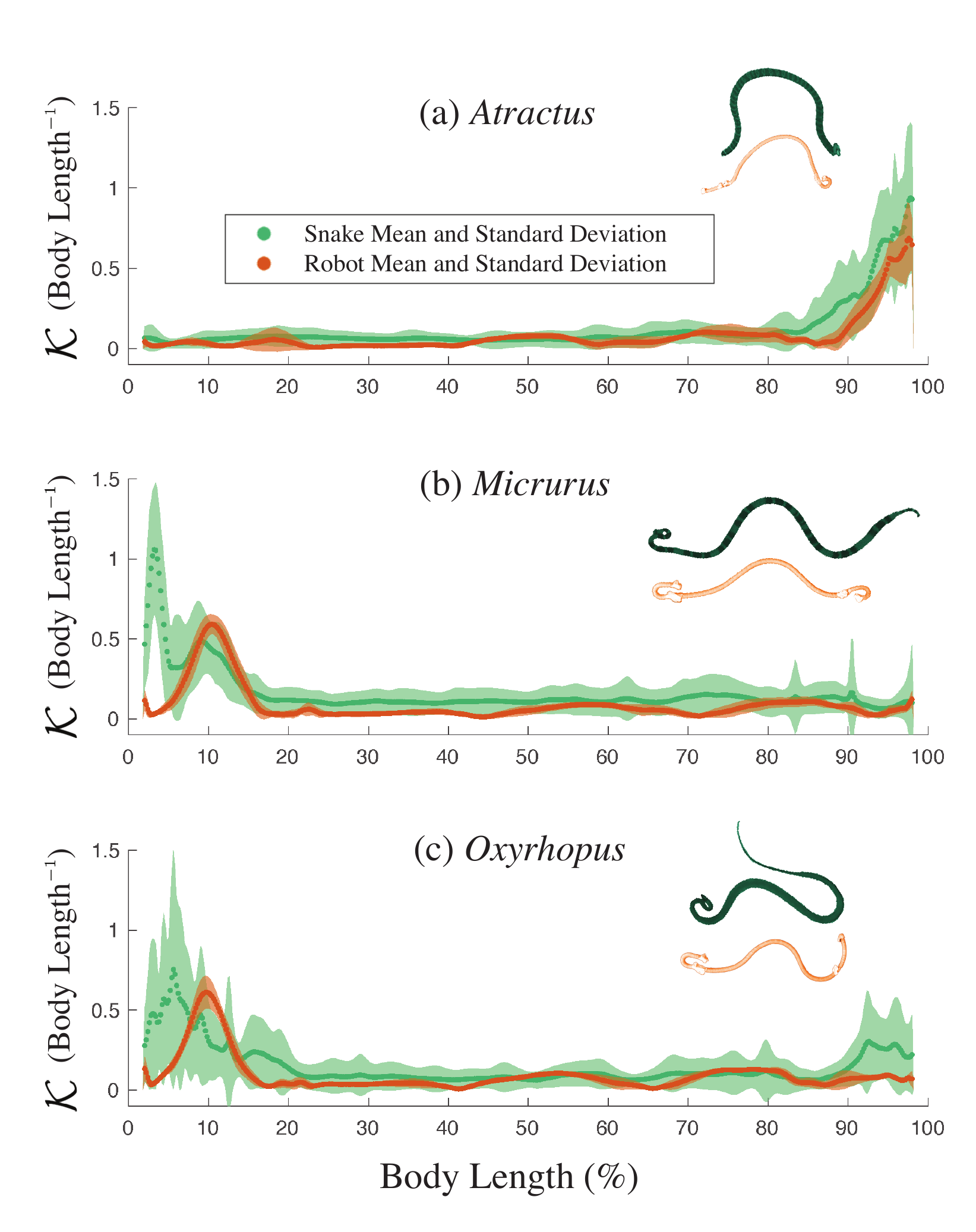}
    \caption{Each unique combination of head, midsection, and tail closely mimics a specific genus in our dataset. One snake (green) and robot (orange) configuration from each genus has been selected for physical comparison.}
    \label{fig:species_snake_comparison}
\end{figure}

The \emph{Atractus} snakes exhibited low curvatures throughout the anterior portions of the body, with high curvatures beginning at approximately 85\% of total body length (Fig. \ref{fig:species_snake_comparison}a).
Although the tails of robots emulating \emph{Atractus} began curving more posteriorly than the snake and did not achieve as high of curvatures during the assembled trials, the robot head and midsection closely match the shapes exhibited by the snake.

As previously observed, \emph{Micrurus} snakes exhibit high curvature towards the anterior $20\%$ of the body, usually with two distinct alternating curves (Fig. \ref{fig:species_snake_comparison}b).
Curvature along the body is much lower, and then begins to increase again at $83\%$, $91\%$, and at $97\%$ near the tight tail coils that were not traced, as they were too tightly coiled.
Robots emulating \emph{Micrurus} were not designed to emulate the standard first lateral head kink close to the anterior portion of the snake head, because it remained kinked throughout the observed trials.
However, the robots were capable of emulating both the location and the magnitude of the second lateral head kink exhibited by the snake.
The robot midsection from approximately $17\%$ to $95\%$ consistently remained within one standard deviation of the mean snake behavior.

The \emph{Oxyrhopus} heads varied more in their number, curvature, and location than the rigorously maintained head kinks of the \emph{Micrurus} (Fig. \ref{fig:species_snake_comparison}c).
\emph{Oxyrhopus} exhibited high curvature at approximately $5\%$ of the body length, which is more posterior than the location of the first \emph{Micrurus} head kink.
Mean curvature was low throughout the midsection of the \emph{Oxyrhopus}, from approximately $20-90\%$ of body length, with some highly variable tail curvature occurring posterior to 90\% of the body.
While \emph{Oxyrhopus} tails would frequently form a loose lateral curve, they were never as tightly coiled as the \emph{Micrurus} or \emph{Atractus} tails.
Robots emulating this genus achieved a similar mean magnitude of curvature, although at a location that is slightly more posterior, due to a closer match with the second lateral head kink of the \emph{Micrurus}.
Despite maintaining a low mean curvature in the tail segment, the robot curvature remained within one standard deviation of the mean snake behavior from approximately $17\%$ throughout the rest of the body.

In no video data does \emph{Oxyrhopus melanogenys} perform a tail coil, nor does \emph{Atractus elaps} perform a lateral neck kink.
Because this study examines the first video dataset collected and first quantitative analysis of anti-predator behaviors for these species, no current video evidence exists to suggest that these species would be capable of full behavioral mimicry.
\section{Conclusion}\label{sec:conclusion}

This work represents the first robotic platform that can be used to reliably emulate the durations and curvatures produced by snakes during anti-predator behavioral displays.
These results demonstrate that although precise control of soft robots is an active field of study, the current technology can be used by behavioral ecologists to test how variations in behavior are perceived by predators.
It has previously been difficult for behavioral ecologists to test how color pattern and behavior contribute to overall predator deterrence because these traits covary in nature \cite{brodie1992}. 
By emulating the anti-predator behaviors of specific genera with soft-robotic devices that can be custom-painted, it is now possible to decouple these traits with robots that are safe for interaction with live predators.
Thus, soft-robotic snakes are a key tool that enable behavioral ecologists to better understand the evolution of anti-predator signaling and mimicry.

\section{Appendix}
All field methods were approved by the University of Michigan Institutional Animal Care and Use Committee (Protocols \#PRO00006234 and \#PRO00008306) and the 
Servicio Nacional Forestal y de Fauna 
Silvestre (SERFOR) in Peru (permit numbers: 
029-2016-SERFOR-DGGSPFFS, 
405-2016-SERFOR-DGGSPFFS, 
116-2017-SERFOR-DGGSPFFS). 
The snakes examined for this study correspond to University of Michigan Museum of Zoology
specimen UMMZ-248449 (\emph{Oxyrhopus melanogenys}, RAB 3557),
UMMZ-248452 (\emph{Micrurus lemniscatus}, RAB 3485),
and Museo de Historio Natural in Lima, Peru specimen 
MUSM-39784 (Atractus elaps, RAB 3405).
\renewcommand{\bibfont}{\normalfont\small}
{\renewcommand{\markboth}[2]{}
\printbibliography}

\end{document}